\documentclass[10pt,twocolumn,letterpaper]{article}

\usepackage{cvpr}
\usepackage{times}
\usepackage{epsfig}
\usepackage{subcaption}
\usepackage{graphicx}
\usepackage{amsmath}
\usepackage{amssymb}
\usepackage{multirow}
\usepackage{algorithm} 
\usepackage{algpseudocode} 

\usepackage[breaklinks=true,bookmarks=false]{hyperref}
\makeatletter
\def\blfootnote{\xdef\@thefnmark{}\@footnotetext}
\makeatother
\cvprfinalcopy 

\def\cvprPaperID{5667} 

\begin{document}

\title{Tracking by Instance Detection: A Meta-Learning Approach}

\author{
Guangting Wang$^{1}$ \qquad Chong Luo$^{2}$ \qquad Xiaoyan Sun$^{2}$ \qquad Zhiwei Xiong$^{1}$ \qquad Wenjun Zeng$^{2}$ \\
University of Science and Technology of China$^{1}$ \qquad Microsoft Research Asia$^{2}$ \\
{\tt\small wgting96@gmail.com\quad \{cluo, xysun, wezeng\}@microsoft.com\quad zwxiong@ustc.edu.cn}}

\maketitle
\thispagestyle{empty}
\begin{abstract}
   We consider the tracking problem as a special type of object detection problem, which we call instance detection. With proper initialization, a detector can be quickly converted into a tracker by learning the new instance from a single image. We find that model-agnostic meta-learning (MAML) offers a strategy to initialize the detector that satisfies our needs. We propose a principled three-step approach to build a high-performance tracker. First, pick any modern object detector trained with gradient descent. Second, conduct offline training (or initialization) with MAML. Third, perform domain adaptation using the initial frame. We follow this procedure to build two trackers, named Retina-MAML and FCOS-MAML, based on two modern detectors RetinaNet and FCOS. Evaluations on four benchmarks show that both trackers are competitive against state-of-the-art trackers. On OTB-100, Retina-MAML achieves the highest ever AUC of 0.712. On TrackingNet, FCOS-MAML ranks the first on the leader board with an AUC of 0.757 and the normalized precision of 0.822. Both trackers run in real-time at 40 FPS. 
   
\end{abstract}
\vspace{-4mm}

\blfootnote{This work was done while Guangting was an intern with MSRA.}

\section{Introduction}

Given a bounding box defining the target object in the initial frame, the goal of visual object tracking is to automatically determine the location and extent of the object in every frame that follows. The tracking problem is closely related to the detection problem, and it even can be treated as a special type of object detection, which we call \emph{instance detection}. The major difference is that object detection locates objects of some predefined classes and its output does not differentiate between intra-class instances. But object tracking only looks for a particular instance, which may belong to any known or unknown object class, that is specified in the initial frame. 

Given the similarity between the two tasks, some object detection techniques are used extensively in object tracking. For example, the region proposal network (RPN), which was proposed in Faster R-CNN detector \cite{FasterRCNN}, has been adopted in SiamRPN tracker and its variants \cite{SiamRPN, SiamRPNpp, DaSiamRPN}. The introduction of multi-aspect-ratio anchors solves the box estimation problem that has been plaguing previous trackers. It has greatly improved the performance of siamese-network-based trackers. More recently, the IoU network \cite{IoUNet}, which is again an innovation in object detection, is applied to object tracking by ATOM and DiMP \cite{ATOM, DiMP} and demonstrates powerful capabilities. 

\begin{figure}[t!]
    \centering
    \includegraphics[width=\linewidth]{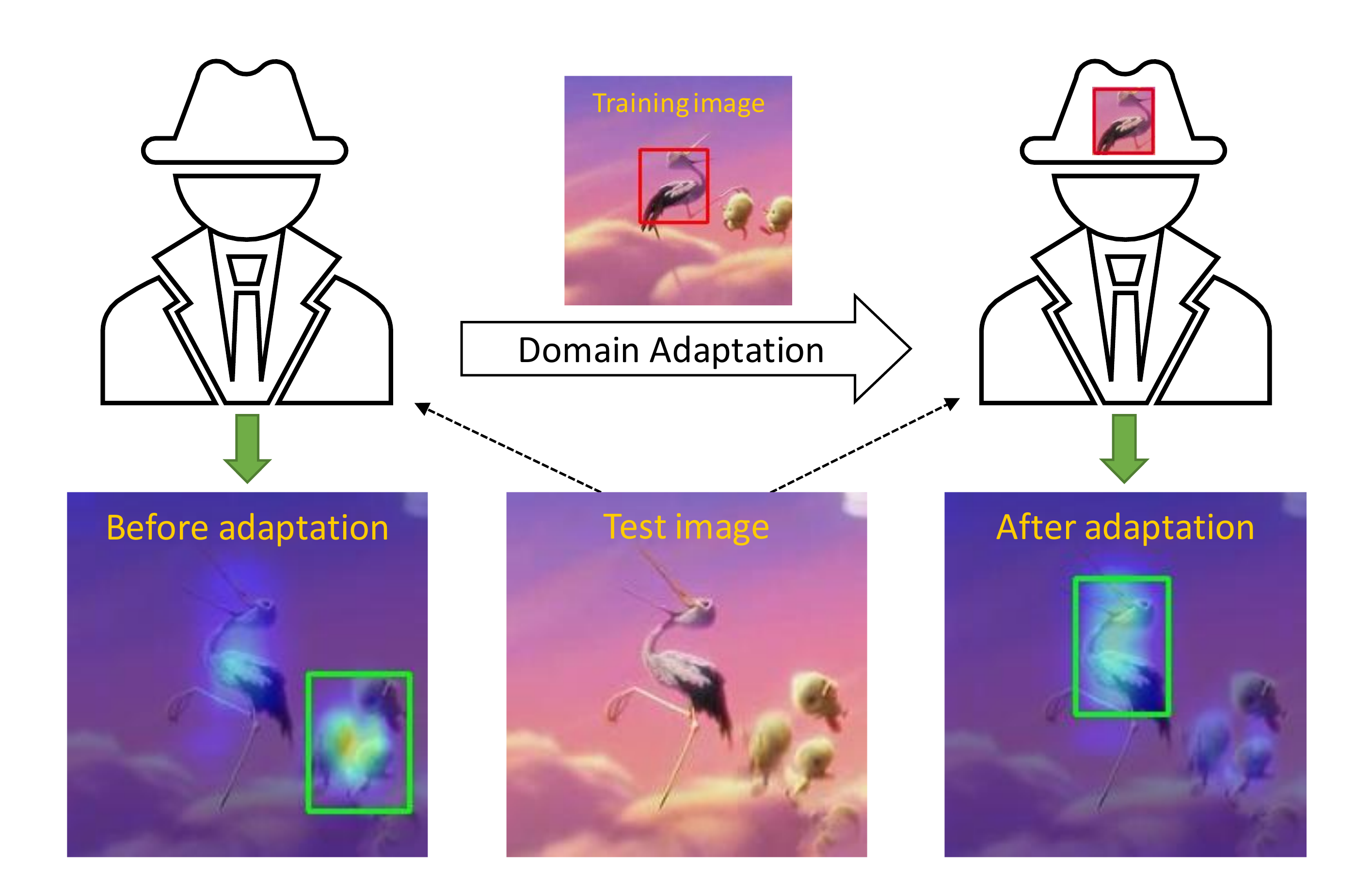}
    \vspace{-8mm}
    \caption{MAML provides an effective way to initialize an instance detector. With a single training image, the detector can be quickly adapted to the new domain (instance). It is capable of locating target object in subsequent frames even when the object has significant appearance changes. }
    \vspace{-4mm}
    \label{FigExample}
\end{figure}

In addition to these approaches that borrow advanced components from object detection to assemble a tracker, we believe that another option is to directly convert a modern object detector into a high-performance tracker. This will allow the tracker to retain not only the advanced components but also the overall design of the base detector. The main challenge is how to obtain a good initialization of the detector so that once a new instance is given, it can efficiently infuse the instance information into the network without overfitting. Fig.\ref{FigExample} illustrates the idea. The detector may behave like a general object detector before adaptation. But after domain adaptation with a single training image, it is able to ``memorize" the target and correctly locate the target in subsequent frames. A recent work by Huang et al. \cite{UnifiedDet} shares a similar vision with us but they still treat tracking as a two-step task, namely class-level object detection and instance-level classification. In the first sub-task, a template image is involved and a separate branch is employed to process the template. 

In this work, we are looking for a neat solution to realize our idea. The constructed tracker will just look like a normal detector, without additional branches or any other modifications to the network architecture. We find that model-agnostic meta learning (MAML) \cite{MAML} offers a learning strategy with which a detector can be initialized as we desire. Based on MAML, we propose a three-step procedure to convert any modern detector into a high-performance tracker. First, pick any detector which is trained with gradient descent. Second, use MAML to train the detector on a large number of tracking sequences. Third, when the initial frame of a test sequence is given, fine-tune the detector with a few steps of gradient descent. A decent tracker can be obtained after this domain adaptation step. During tracking, when new appearances of the target are collected, the detector can be trained with more samples to achieve an even better adaptation capability. 

Following the proposed procedure, we build two instance detectors, named Retina-MAML and FCOS-MAML, based on advanced object detectors RetinaNet \cite{RetinaNet} and FCOS \cite{FCOS}. During offline training, we further introduce a kernel-wise learnable learning rate in MAML to improve the expressive ability of gradient based updating. Evaluations of the trackers are carried out on four major benchmarks, including OTB, VOT, TrackingNet and LaSOT. 
System comparisons show that both trackers achieve competitive performance against state-of-the-art (SOTA) trackers. On OTB-100, Retina-MAML and FCOS-MAML appear to be the best-performing trackers, with AUCs of 0.712 and 0.704, respectively. Retina-MAML achieves an EAO of 0.452 on VOT-2018. FCOS-MAML achieves an AUC of 0.757 on TrackingNet, ranking number one on the leader board. Additionally, both trackers run in real-time at 40 FPS.

\section{Related Work} \label{SectionRelatedWork}

\subsection{CNN-based visual object tracking}

With the great success of deep learning and convolutional neural networks (CNN) in various computer vision tasks, there emerge an increasing number of CNN-based trackers. We divide CNN-based trackers into two categories, depending on whether an explicit template is used. 

Most siamese-network-based trackers \cite{SiamFC, SiamRPN, SiamRPNpp, SPM} fall into the first category, which we call \emph{template-based} methods. The target appearance information is stored in an explicit template. In SiamFC \cite{SiamFC}, features are extracted from the template and the search region using the same offline-trained CNN. A cross-correlation operation is then adopted to compute the matching scores. A main drawback of SiamFC is that it only evaluates the candidates with the same shape as the initial box. SiamRPN \cite{SiamRPN} solves this problem by borrowing the RPN idea from object detectors. Later, SPM-Tracker \cite{SPM} borrows the architecture from two-stage detectors and achieves an improved performance. Currently, the best-performing trackers in this category are ATOM \cite{ATOM} and DiMP \cite{DiMP}, which leverage the most advanced IoUNet \cite{IoUNet} for precise object localization. 

Template-based methods usually run very fast, because the CNN used to extract features does not need to be online updated. However, as the tracking proceeds, new target appearances should be integrated into the template for a better performance. But most methods lack an effective model for template online updating. 
This limitation has created a performance ceiling for template-based trackers. 

The other category is \emph{template-free} methods \cite{T-CNN, MDNet, RT-MDNet}, which intend to store the target appearance information within the neural network, in the form of fine-tuned parameters. The challenge in designing template-free trackers is how to quickly infuse the instance information to the network without overfitting. MDNet \cite{MDNet} divides the CNN into shared layers and domain-specific layers. The shared layers provide a reasonable initialization and the domain-specific layers are online trained with the new instance. Due to the limitation of the conventional training strategy, MDNet takes many iterations to converge and fewer iterations cause serious performance degradation. As a result, MDNet is too slow to be used in real-time scenarios. 

We find that template-free trackers are neat solutions. They do not need to maintain an external template and the network architecture looks just like a detector. Domain adaptation and online update can be achieved by a unified online training procedure. However, it is still quite challenging to achieve a good performance-speed tradeoff for this type of trackers. 

\begin{figure*}[t]
    \centering
    \includegraphics[width=1.0\linewidth]{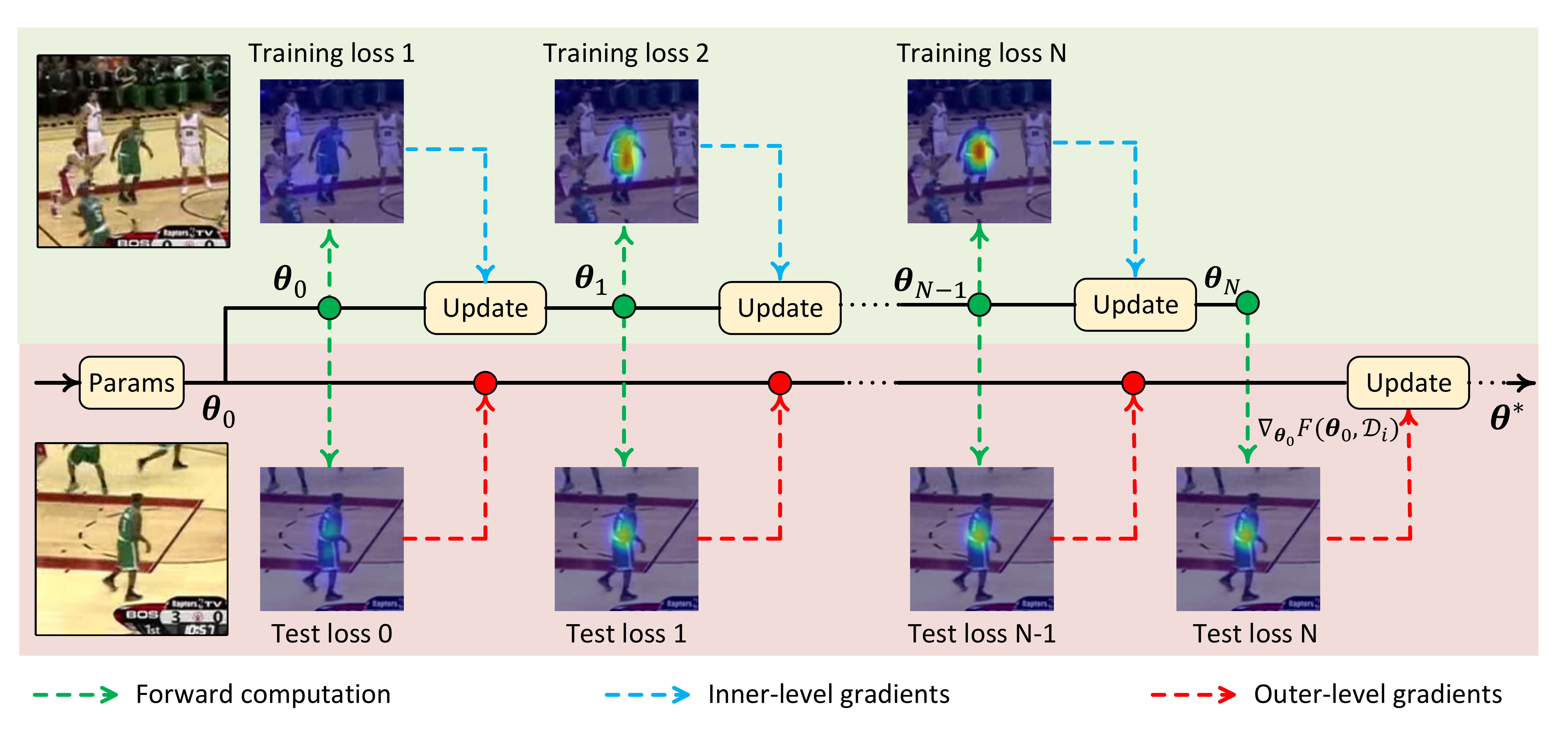}
    \vspace{-10mm}
    \caption{Illustration of our training pipeline. The first row is the inner training loop. A few steps of SGD optimization is performed on the support images. The updated parameters after each step are used for calculating the meta-gradient based on testing images. Best viewed in colors.}
    \label{FigArchitecture}
\end{figure*}

\subsection{Meta learning and its application to tracking}

The goal of meta-learning is to train a model on a variety of learning tasks, such that it can solve new learning tasks using only a small number of training samples \cite{MAML}. When we view object tracking as an instance detection task, the tracker is trained on a variety of instance detection tasks so that it can quickly learn how to detect a new instance using only one or a few training samples from the initial or previous frames. We find that the tracking task is a perfect example to apply meta-learning. 

Model-agnostic meta-learning (MAML) \cite{MAML} is an important algorithm for meta learning. It helps the network to learn a set of good initialization parameters that are suitable for fine-tuning. During training, the parameters of the model are explicitly trained such that a small number of gradient steps with a small amount of training data from a new task will produce good generalization performance on that task. The most striking merit of MAML is that it is compatible with any model trained with gradient descent and applicable to a variety of different learning problems. Because of this, MAML is a perfect candidate to realize our idea, which is to convert any advanced object detectors (trained with gradient descent) into a tracker. Later, MAML++ \cite{MAML++} introduces a set of tricks to stabilize the training of MAML. MetaSGD \cite{MetaSGD} proposes to train learnable learning rates for every parameter. 
In the area of object tracking, Meta-Tracker \cite{MetaTracker} is the first to use MAML for the domain adaptation step of MDNet. MetaRTT \cite{MetaRTT} further applies MAML for the online updating step. Basically, their main purpose is to accelerate the online training of existing trackers, including MDNet \cite{MDNet}, CREST \cite{CREST} and RT-MDNet \cite{RT-MDNet}. 
We argue that, since meta learning provides a mechanism to quickly adapt a deep network to model a particular object and avoid overfitting, why not directly convert a modern object detector into a tracker, instead of making a slow tracker faster? Huang et al. \cite{UnifiedDet} have the same idea. They propose to learn a meta layer in detection head by MAML. However, they still introduce a template in the first part of the tracker called class-level object detection. The complex design results in a slow speed.

\section{Learning an Instance Detector with MAML} \label{SectionApproach}

The key to convert a detector into an instance detector (a tracker) is to provide a good initialization of the detector, so that it can quickly adapt to a new instance when only the initial frame is available. In this section, we present the approach to learn an instance detector with MAML. The complete steps to construct a tracker will be detailed in the next section. The training data in this learning step are videos with ground-truth labeling of the target object on each frame. 

Formally, given a video $V_i$, we collect a set of training samples, denoted by $\mathcal{D}^{s}_{i}$. It is also called the \emph{support set} in meta learning. A detector model is defined as $h(x; \boldsymbol{\theta}_0)$, where $x$ is the input image and $\boldsymbol{\theta}_0$ is the parameters of the detector. We update the detector on the support set by a $k$-step gradient descent (GD) algorithm:
\begin{equation} \label{EqInner}
    \begin{split}
        \boldsymbol{\theta}_k &\equiv \text{GD}_{k}(\boldsymbol{\theta}_0, \mathcal{D}^{s}_{i}) \quad,\text{ and} \\
        \boldsymbol{\theta}_{k} &= \boldsymbol{\theta}_{k-1} - \alpha \frac{1}{|\mathcal{D}^{s}_{i}|}\sum_{(x, y) \in \mathcal{D}^{s}_{i}}\nabla_{\boldsymbol{\theta}_{k-1}} \mathcal{L}(h(x; \boldsymbol{\theta}_{k-1}),y),
    \end{split}
\end{equation}
where $\mathcal{L}$ is the loss function and $(x, y)$ is a data-label pair in the support set. The procedure in Eqn. (\ref{EqInner}) is called \textit{inner-level optimization}. To evaluate the generalization ability of the trained detector, we collect another set of samples $\mathcal{D}^{t}_{i}$ from the same video $V_i$ and they are called the \emph{target set}. We calculate the loss on the target set by applying the trained detector, which can be written as:
\begin{equation} \label{EqMAMLLoss}
    \begin{split}
        F(\boldsymbol{\theta}_{0}, \mathcal{D}_i) &= \frac{1}{|\mathcal{D}^{t}_{i}|}\sum_{(x, y) \in \mathcal{D}^{t}_{i}}\mathcal{L}(h(x; \boldsymbol{\theta}_k), y) \\
    \end{split}
\end{equation}
\noindent where $\mathcal{D}_i = \{\mathcal{D}^{s}_i, \mathcal{D}^{t}_i\}$ denotes the combined support set and target set. The overall training objective is to find a good initialization status $\boldsymbol{\theta}_0$ for any tracking video. It can be formulated as:
\begin{equation} \label{EqMAMLOpt}
    \begin{split}
        \boldsymbol{\theta}^{*} = \arg\min_{\boldsymbol{\theta}_0} \frac{1}{N}\sum_{i}^{N}F(\boldsymbol{\theta}_0, \mathcal{D}_{i}), 
    \end{split}
\end{equation}
where $N$ is the total number of videos. The procedure in Eqn. (\ref{EqMAMLOpt}) is called the \textit{outer-level optimization}, which can be solved by gradient-based methods like Adam \cite{Adam}. The outer-level gradients are back-propagated through the inner-level computational graph. The only assumption about the detector $h$ is that it is differentiable. Therefore, this approach is readily applicable to most deep learning based detectors.

Fig. \ref{FigArchitecture} illustrates this training pipeline. In the training phase, we only sample a pair of images from the dataset. Following the practice in DaSiamRPN \cite{DaSiamRPN}, these two images may come from either the same sequence or different sequences. The first image will be zoomed in/out by a constant factor (1.08 in our experiments) so that a support set with three images is constructed for the inner-level optimization. The second image is viewed as the target set with single image for calculating the outer-level loss. We use a 4-step GD for the inner-level optimization and Adam solver \cite{Adam} for the outer-level optimization. To stabilize the training and strengthen the power of detector, we make the following modifications to the original MAML algorithm.

\textbf{Multi-step loss optimization.} MAML++ \cite{MAML++} proposes to take the parameters after \textit{every} step of inner-level GD to minimize the loss on target set, instead of only using the parameters after \textit{final} step. Mathematically, Eqn. (\ref{EqMAMLLoss}) can be re-written into:
\begin{equation}
    \begin{split}
        F(\boldsymbol{\theta_0}, \mathcal{D}_i) &= \frac{1}{|\mathcal{D}^{t}_i|}\sum_{(x, y) \in \mathcal{D}^{t}_i} \sum_{k=0}^{K}\gamma_k\mathcal{L}(h(x; \boldsymbol{\theta}_k), y),\\
    \end{split}
\end{equation}
where $K$ is the number of inner-level steps and $\gamma_k$ is the loss weight for each step. Note that our formulation is slightly different from that in MAML++. The initialization parameter $\boldsymbol{\theta}_0$ (before updating) also contributes to the outer-level loss. In our experiments, we find this trick is crucial for stabilizing the gradients.

\textbf{Kernel-wise learnable learning rate.} In standard MAML, the learning rate $\alpha$ in the inner-level optimization is a predefined constant. MetaSGD \cite{MetaSGD} proposes to specify a learnable learning rate for each parameter in the model. Therefore, the GD algorithm in Eqn. (\ref{EqInner}) can be re-written into:
\begin{equation} \label{EqKLL}
    \begin{split}
        \boldsymbol{\theta}_{k+1} = \boldsymbol{\theta}_{k} - \boldsymbol{\alpha}\odot\frac{1}{|\mathcal{D}^{s}_{i}|}\sum_{(x, y) \in \mathcal{D}^{s}_{i}}\nabla_{\boldsymbol{\theta}_{k}} \mathcal{L}(h(x; \boldsymbol{\theta}_{k}), y),
    \end{split}
\end{equation}
where $\boldsymbol{\alpha}$ is a tensor which has the same size as $\boldsymbol{\theta}_k$. Notation $\odot$ denotes the element-wise product. However, setting up a learning rate for every parameter will double the model size. In contrast, we arrange the learnable learning rates in a kernel-wise manner. Specifically, for a convolution layer with $C_{out}$ output channels, we define a learning rate for each convolutional kernel and this only introduces an additional number of $C_{out}$ learnable parameters, which are negligible in the model.

\section{Retina-MAML and FCOS-MAML} \label{SectionImpDetails}

This section provides the details of the proposed three-step procedure to build a tracker. Specifically, we will present detector choices, offline training details, and the online tracking process for two trackers named Retina-MAML and FCOS-MAML. 

\subsection{Detectors} \label{SectionOurDetector}

As MAML is a model-agnostic learning approach, we are free to choose any modern detector trained with gradient descent as the base to build a tracker. As the first attempt in this direction, we choose two single-stage detectors which run faster and are fairly easy to manipulate than their two-stage counterparts. However, we do not see any obstacles in using two-stage detectors in our approach. 

Single-stage detectors are usually composed of a backbone network and two heads, namely classification head and regression head. The backbone network generates feature maps for the input image. Based on the feature maps, the objects are scored and localized. 

RetinaNet \cite{RetinaNet} is a representative single-stage object detector. Each pixel in the feature maps is associated with several predefined prior boxes, or anchors. The classification head is trained to classify whether each anchor has a sufficient overlap with an object. The regression head is trained to predict the relative differences between each anchor and the corresponding ground-truth box. Similar design can be found in many existing detectors, which are grouped into a family of anchor-based detectors. 

\begin{figure}[t!]
    \centering
    \includegraphics[width=1.0\linewidth]{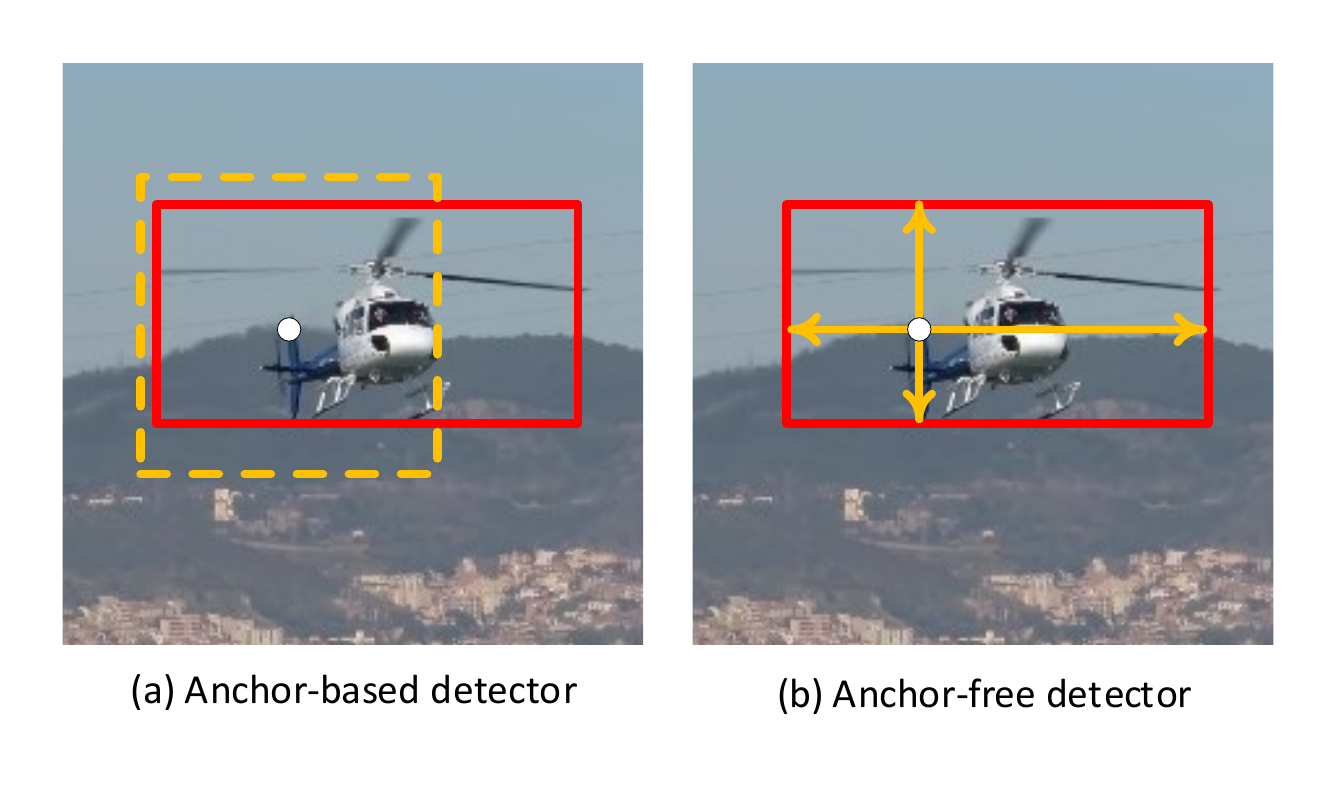}
    \caption{(a) Anchor-based detectors predict the relative differences between the anchor and the ground-truth box. The dotted yellow box represents an anchor. (b) Anchor-free detectors directly estimate four offsets from the pixel to object boundaries.}
    \label{FigEDetectorAFAB}
\end{figure}

Recently, the concept of anchor-free detection has received a lot of attention. As the name suggests, no anchor is defined. FCOS \cite{FCOS} is a representative detector in this category. After the backbone network generates the feature maps, the classification head is trained to classify whether each pixel in the feature maps is within the central area of an object. Meanwhile, the regression head directly estimates the four offsets from the pixel to the object boundaries. Fig. \ref{FigEDetectorAFAB} depicts the core design difference between anchor-free and anchor-based detectors.

Next, we make some simplifications to the chosen detectors RetinaNet and FCOS. These simplifications improve the tracker's speed but will not affect the tracking performance. We believe so because visual object tracking is performed frame-by-frame on a video sequence. Subsequent video frames have strong temporal correlations, so the location and extent of the target object in the previous frame provide a close estimate of those in the current frame. Usually, tracking is performed on a square-shaped search region, which is further scaled to a fixed size before being passed to the tracking network. From the tracker's point of view, the size distribution of target object is very concentrated. Therefore, it is not necessary to use the FPN module, which is mainly adopted to handle large scale variations, in RetinaNet and FCOS. Additionally, the vanilla-version of FCOS uses three network heads, one common regression head and two centerness/classification heads. Since tracking only needs to classify target and non-target, we only keep the centerness branch to produce classification scores. 

The second step is to initialize a detector with offline MAML training. As the detailed algorithm has been introduced in the previous section, we provide implementation details here. 

\begin{figure}[t]
    \centering
    \includegraphics[width=0.9\linewidth]{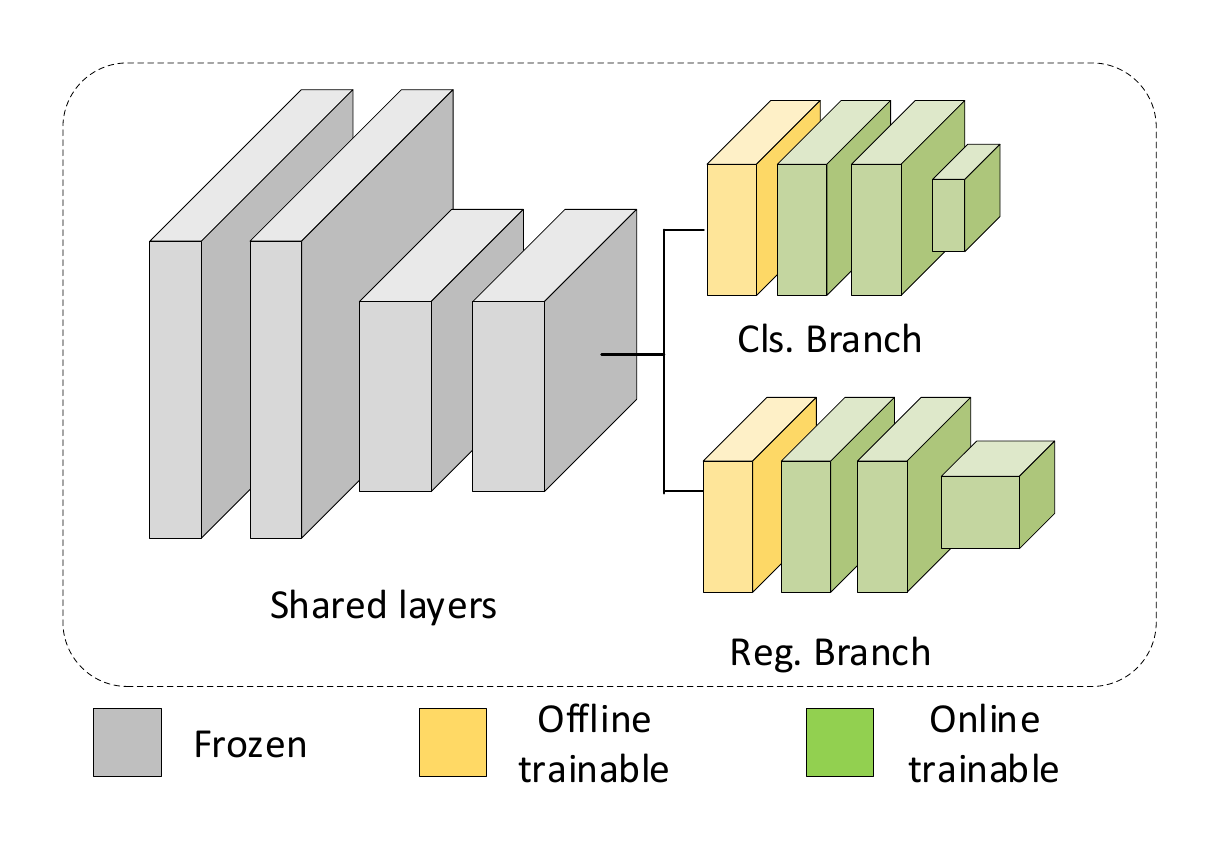}
    \vspace{-4mm}
    \caption{We adopt ResNet-18 as the backbone. The first three blocks are frozen after ImageNet pre-training and block-5 is removed. Block-4 is independently trained in the classification branch and the regression branch during offline training. Online training only involves a subset of trainable layers.}
    \label{FigBranch}
\end{figure}

\textbf{Network architecture.} Fig. \ref{FigBranch} depicts the detection network we use for MAML training. In both detectors, the CNN backbone used for feature extraction is ResNet-18. The parameters in the first three blocks are pre-trained with ImageNet and frozen during offline training. The last block (block-5) is discarded so that the stride to output feature maps is 8. We make two independent copies of block-4 and put them to the respective branches. This is not a necessary treatment for our approach to work, just to allow us to analyze the effect of online updating during tracking. 
For RetinaNet, we pre-define a single anchor box with a size of $64\times64$ pixels. In our experiments we find that this single-anchor setting performs slightly better than the multi-anchor setting in SiamRPN \cite{SiamRPN}. 

\subsection{Offline MAML training}

\textbf{Loss definition.} For Retina-MAML, an anchor box is assigned a positive (or negative) label when its intersection-over-union (IoU) overlap with the ground-truth box is greater than 0.5 (or less than 0.3). We use focal loss and smooth L1 loss to train the classification branch and regression branch, respectively. For FCOS-MAML, we adopt L2 loss to supervise the training of centerness scores. The loss function in regression branch is L1 loss. 

\textbf{Training data.} Following other modern trackers \cite{ATOM, DiMP}, we use four datasets for offline training, namely MS-COCO \cite{COCO}, GOT10k \cite{GOT10K}, TrackingNet \cite{TrackingNet} and LaSOT-train \cite{LaSOT}. In LaSOT and TrackingNet, we only sample one frame for every three or ten frames. The training images are cropped and resized into a resolution of $263 \times 263$. Standard data augmentation mechanisms like random scaling and shifting are adopted.

\textbf{Optimization.} As noted in Section \ref{SectionApproach}, we use 4-step GD for inner-level optimization during offline training period. The kernel-wise learnable learning rate $\boldsymbol{\alpha}$ is initialized to 0.001. The multi-step loss weights $\gamma_k$ are initialized as equal contribution and gradually anneal to (0.05, 0.10, 0.2, 0.30, 0.35), giving more weight and attention to later steps.  
For the outer-level optimization, we adopt Adam optimizer \cite{Adam} with a starting learning rate 0.0001. In each iteration, 32 pairs of images are sampled. The detector is trained for 20 epochs, with 10,000 iterations per epoch. To accelerate the training, we use first-order approximation \cite{MAML++} in the first 15 epochs.

\subsection{Online training and tracking} \label{SectionInference}

\begin{algorithm}[ht!]
	\caption{Online tracking algorithm}
	\label{AlgTracking}
	\small
    \textbf{Input:} Frame sequence $\{I_i\}_{i=1}^{N}$, detector $h(\cdot;\boldsymbol{\theta})$, initialization bounding box $B_1$, update interval $u$. \\
    \textbf{Output:} Tracking results $\{B_i\}_{i=1}^{N}$
	\begin{algorithmic}[1]
	    \State Generate search region image. $S_1\leftarrow\text{SR}(I_1, B_1)$  
	    \State Initialize the support set. $\mathcal{D}^s\leftarrow\{\text{DataAug}(S_1)\}$ 
	    \State Model update in Eqn. (\ref{EqInner}). $\boldsymbol{\theta}\leftarrow\text{GD}_{5}(\boldsymbol{\theta}, \mathcal{D}^s)$
		\For {$i=2,...,N$}
		\State Detect objects represented in bounding box and score. $\{B_{\text{det}}^{j}, c_j\}_{j=1}^{M}\leftarrow h(\text{SR}(I_i, B_{i-1});\boldsymbol{\theta})$
		\If {all $c_j < 0.1$}
		\State $B_{i}\leftarrow B_{i-1}$
		\State \textbf{continue}
		\EndIf
		\State Add penalties and window priors to $\{B^{j}_{\text{det}}, c_j\}_{j=1}^{M}$
		\State Select the box with the highest score $c^*$. $B_i\leftarrow B^{*}_{det}$
		\State Linear interpolate shape. $B_i\leftarrow \text{Inter}(B_i, B_{i-1})$
		\State Update the support set $\mathcal{D}^s$.
		\If {$i\mod u = 0$ \textbf{or} distractor detected}
		    \State Model update in Eqn. (\ref{EqInner}). $\boldsymbol{\theta}\leftarrow\text{GD}_1(\boldsymbol{\theta}, \mathcal{D}^s)$
		\EndIf
		\EndFor
	\end{algorithmic} 
\end{algorithm} 

The third step is \emph{domain adaptation} when a new video sequence is given. In the initial frame, the instance to be tracked is indicated by a ground-truth bounding box. We generate a patch with resolution $263 \times 263$ according to the given bounding box. As with the offline training, we also adopt zoom in/out data augmentation to construct the support set. The tracker is updated by a 5-step GD as described in Eqn. (\ref{EqKLL}). 

After domain adaptation, the detector is now capable of tracking the target object in subsequent frames. For each search region patch, the detector locates hundreds of candidate bounding boxes, which are then passed to a standard post-processing pipeline as suggested in SiamRPN \cite{SiamRPN}. Specifically, shape penalty function and cosine window function are applied to each candidate. Finally, the candidate box with the highest score is selected as the tracking result and its shape is smoothed by a linear interpolation with the result in the previous frame. 

During tracking, the support set is gradually enlarged. The tracker can be online trained at a pre-defined interval based on the updated support set. This process is often called \emph{online updating} in tracking. If a tracking result has a score above a predefined threshold, it will be added into the support set. We buffer at most 30 training images in the support set. Earlier samples, except the initial one, will be discarded when the number of images exceeds the limit. After every $n$ frames ($n=10$ in our implementation) or when a distracting peak is detected (when the peak-to-sidelobe is greater than 0.7), we perform online updating. In this case, we only use 1-step GD to maintain a high tracking speed. On average, our tracker can run at 40 FPS on a single NVIDIA P100 GPU card. The online tracking procedure is summarized in Alg. \ref{AlgTracking}. 

\begin{table}[]
\centering
\small
\setlength\tabcolsep{2.5pt}
\begin{tabular}{c|c|c|c|c|c}
\hline
\multirow{2}{*}{Detector}&Domain                   & OTB-100 & VOT-18 & LaSOT & TrackingNet \\ 
                        &Adaptation                & (AUC)  & (EAO)  & (AUC) & (AUC) \\ \hline
\multirow{2}{*}{Baseline} & before                   & 0.460  & 0.137  & 0.391 & 0.601 \\ 
                        & after                    & 0.487  & 0.174 &  0.391 & 0.634\\ \hline
\multirow{2}{*}{MAML}   & before                   & 0.464  & 0.162  & 0.387 & 0.626 \\ 
                        & after                    & \textbf{0.671}  & \textbf{0.341} & \textbf{0.511} & \textbf{0.743} \\ \hline
\end{tabular}
\caption{MAML training allows a detector to quickly adapt to a new domain, and therefore is the key in turning a detector into a tracker.}
\label{TableCMPNaive}
\end{table}

\begin{figure}
\centering
\begin{subfigure}{1.0\linewidth}
  \centering
  \includegraphics[width=.90\linewidth]{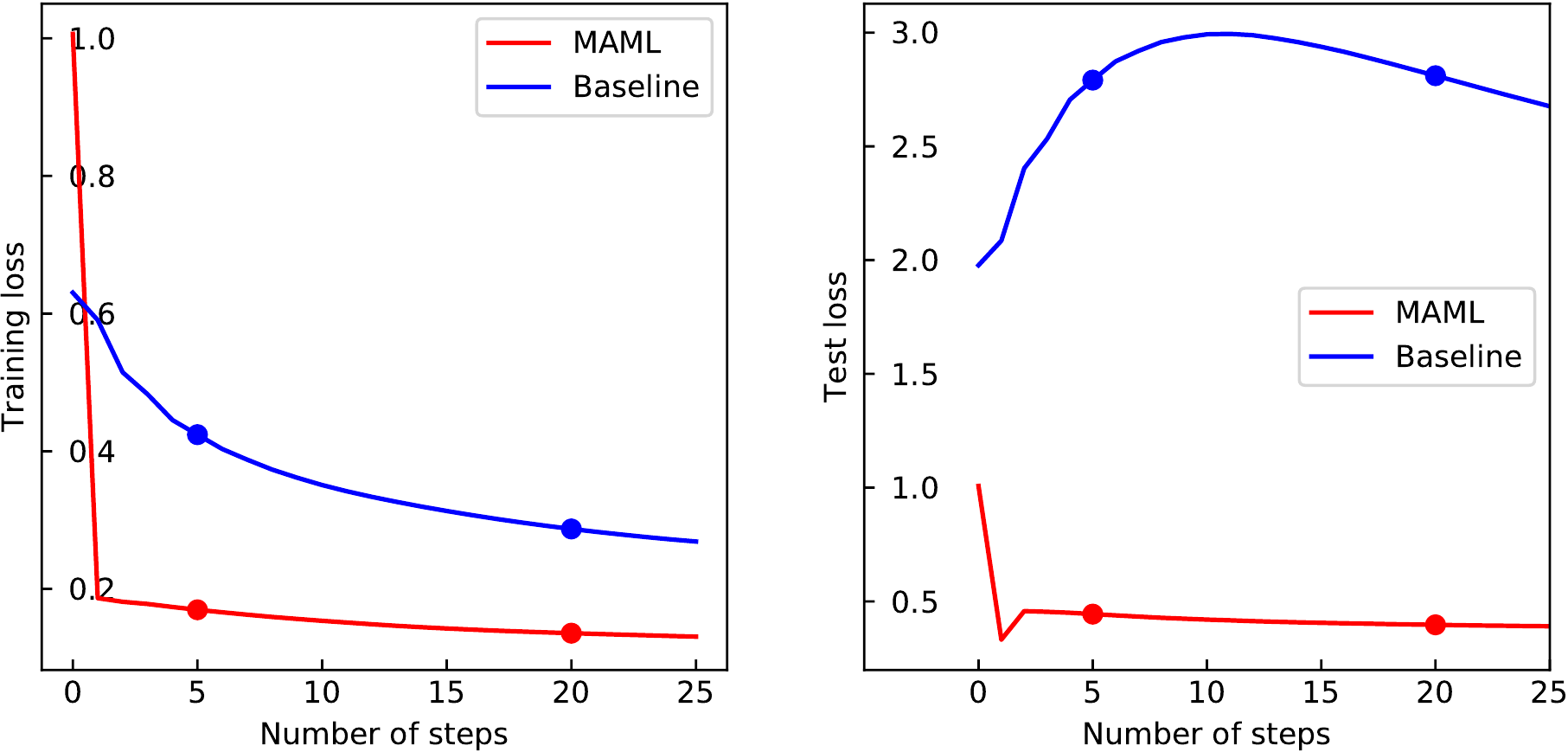}
  \caption{Loss curve}
  \label{fig:sub1}
\end{subfigure}%
\hfill
\begin{subfigure}{1.0\linewidth}
  \centering
  \includegraphics[width=1.0\linewidth]{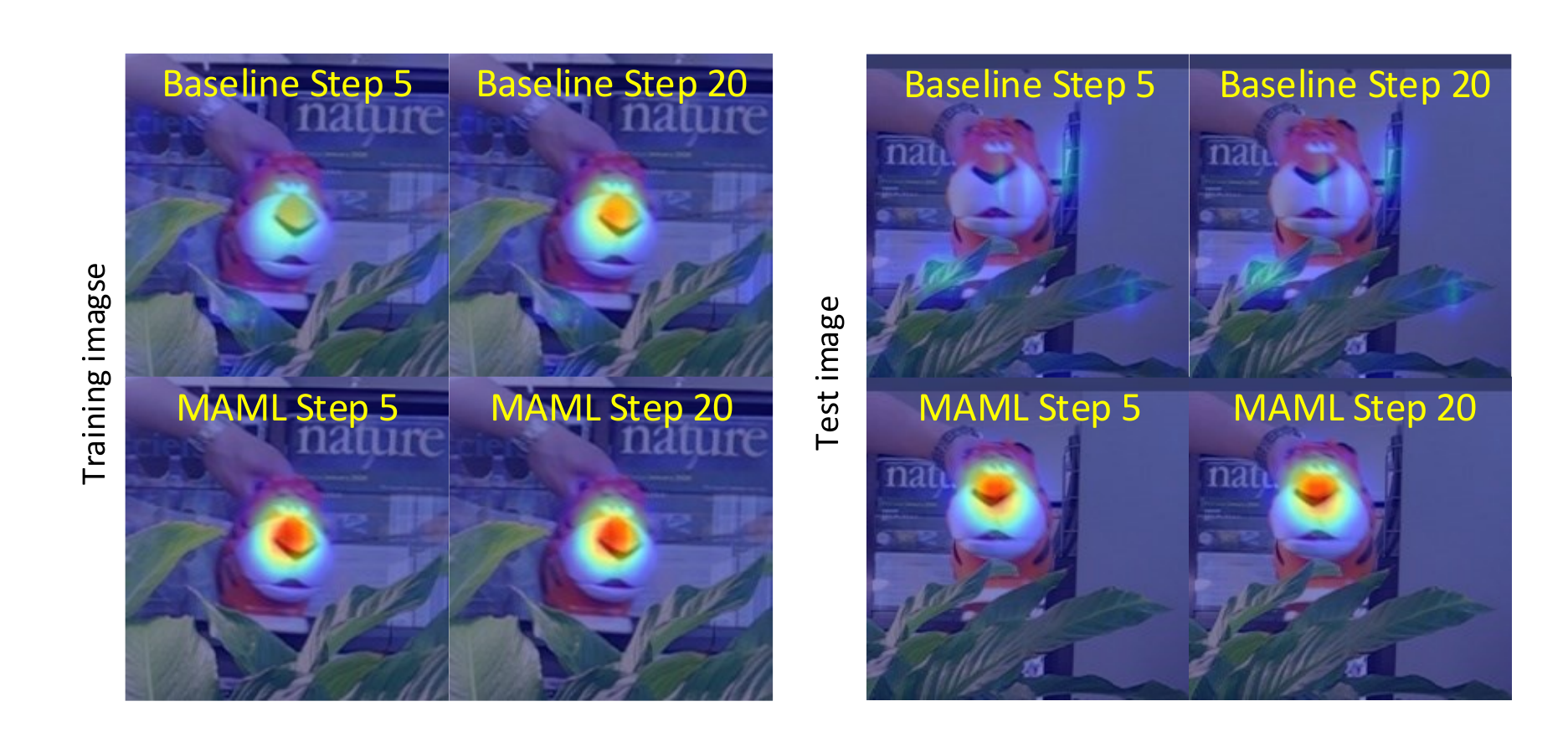}
  \vspace{-8mm}
  \caption{Visualization}
  \label{fig:sub2}
\end{subfigure}
\vspace{-4mm}
\caption{Comparison of the MAML detector and the baseline detector during domain adaptation. (a) Quantitative losses on the training image and a testing image. (b) Visualization of the corresponding score maps. MAML detector convergences quickly and has strong generalization ability.}
\label{FigCMP_MAML_Naive}
\end{figure}

\begin{table}[]
\centering
\setlength\tabcolsep{2.5pt}
\begin{tabular}{|cc|cccc|}
\hline
\multicolumn{2}{|c|}{KLLR in} & OTB-100 & VOT-18 & LaSOT & TrackingNet \\
cls.          & reg.         & (AUC)  & (EAO)  & (AUC)  & (AUC)\\ \hline
              &              & 0.628  & 0.313  & 0.490  & 0.733 \\ 
\checkmark    &              & 0.661  & 0.368  & 0.502 & 0.737 \\ 
              & \checkmark   & 0.676  & 0.315  & 0.504  & 0.744 \\ 
\checkmark    & \checkmark   & \textbf{0.704}  & \textbf{0.392}  & \textbf{0.523} & \textbf{0.757} \\ \hline
\end{tabular}
\caption{Ablation analysis of kernel-wise learnable learning rate. Cls. and reg. denote the classification branch and the regression branch, respectively.}
\label{TableKLL}
\end{table}

\section{Experiments} \label{SectionExp}

\subsection{Ablation study}
Meta-learning is the key in turning a detector into a tracker. In a nutshell, an instance detector can be built by offline MAML training and domain adaptation (online training of the initial frame), and online updating further boosts the performance. In this section, we use FCOS-MAML to carry out the ablation study, which is centered on offline MAML training and online updating. The experiments are conducted on four tracking benchmarks \cite{OTB, VOT18, LaSOT, TrackingNet}, following the official evaluation protocols.

\subsubsection{Offline MAML training}

Without MAML training, one could train a general object detector with standard gradient descent. However, such a detector is not capable of domain adaptation with only a few steps of updating using the samples from the initial frame. 

To demonstrate the importance of MAML training, we offline train the FCOS detector with standard GD and MAML on the same dataset. They are called \emph{baseline detector} and \emph{MAML detector} in this subsection. The performance is presented in Table \ref{TableCMPNaive}. Without domain adaptation, both detectors perform poorly in the tracking task. This is natural because they do not remember any information of the tracking target. However, after domain adaptation with a 5-step GD, the MAML detector shows an clear advantage over the baseline detector. The AUC on OTB-100 is greatly improved from 0.464 to 0.671. In contrast, the baseline detector only slightly benefits from domain adaptation. 

We can get a more intuitive impression of the two detectors from Fig. \ref{FigCMP_MAML_Naive}. Fig. \ref{FigCMP_MAML_Naive}(a) shows the loss curve of the two detectors during domain adaptation. Note that both detectors use the same GD algorithm in this process, but the MAML detector has a much better adaptation capability. 
For the training image, the loss of the MAML detector quickly drops to a small value after only one-step GD updating. The convergence speed of the baseline detector is much slower and the loss is still large after 20 steps of updating. The right of Fig. \ref{FigCMP_MAML_Naive}(a) shows that the loss on the testing image even rises as the training proceeds. Fig. \ref{FigCMP_MAML_Naive}(a) visualizes the response maps on the training and testing images generated by the two detectors. The MAML detector clearly locates the tracking target after 5-step GD in both training and testing images, while the baseline detector does not make any progress even after 20 steps of GD. 

\begin{table}[]
\centering
\setlength\tabcolsep{2.5pt}
\begin{tabular}{|cc|cccc|c|}
\hline
\multicolumn{2}{|c|}{Online} & OTB-100 & VOT-18 & TrackingNet & LaSOT&  Speed \\
cls.             & reg.            & (AUC)  & (EAO)  & (AUC)    & (AUC)   & (FPS) \\ \hline
                  &                 & 0.671  & 0.341  & 0.743  & 0.511          & 85                        \\
\checkmark       &                 & 0.690  & \textbf{0.394}  & 0.747 & \textbf{0.523}   & 58                        \\
\checkmark       & \checkmark               & \textbf{0.704}  & 0.392  & \textbf{0.757} & 0.496 & 42                        \\ \hline

\end{tabular}
\caption{Ablation analysis of the online updating strategy. The baseline tracker without online updating achieves a good performance-speed tradeoff. Online updating both branches is the best choice for tracking short sequences. }
\label{TableUpdate}
\end{table}

\subsubsection{Kernel-wise learnable learning rate}

The model learns information about the target objects from gradients. We propose to use learnable learning rates (KLLR) in a kernel-wise manner. These learning rates guide the directions of gradients and strengthen the power for our model. In this section, we train several FCOS-MAML detectors with or without KLLR. Experimental results in Table \ref{TableKLL} show that the model can benefit from KLLR in both classification branch and regression branch. 

\subsubsection{Online updating strategy}

Our trackers perform two types of online training, one on the initial frame for domain adaptation and the other on the collected samples during tracking. The latter is known as online updating. While domain adaptation is a must-have training procedure for instance detectors, online updating is optional. We first evaluate the simplest baseline, which does not perform online updating at all. Surprisingly, this scheme achieves competitive performance on all the four benchmarks, as shown in Table \ref{TableUpdate}. This version of FCOS-MAML can run very fast at up to 85 FPS. When online updating is adopted, FCOS-MAML achieves increased performance with slightly reduced speed. Comparing the last two rows, we have an interesting finding that is contrary to the conventional wisdom. It was previously believed that online updating the regression branch may harm the tracker's performance due to the aggregated errors. However, our results show that, except for the LaSOT dataset which is composed of very long sequences, online updating both branches seems to be the best choice. 

\begin{table}[]
\small
\setlength{\tabcolsep}{4pt}
\begin{center}
\begin{tabular}{l|ccc|c}
\hline
  \multirow{2}{*}{Tracker} & \multicolumn{3}{c|}{AUC score (OPE)} & Speed  \\
                           & OTB-2013    & OTB-50    & OTB-100   &  (FPS) \\ \hline

 CFNet \cite{CFNet}                  & 0.611       & 0.530     & 0.568     &75    \\
 BACF \cite{BACF}                  & 0.656       & 0.570     & 0.621     &35    \\
 ECO-hc \cite{ECO}                 & 0.652       & 0.592     & 0.643     &60    \\
 MCCT-hc \cite{MCCT}               & 0.664       & -         & 0.642     &45    \\ 
 ECO \cite{ECO}                    & 0.709       & 0.648     & 0.687     &8     \\
 RTINet \cite{RTINet}                 & -           & 0.637     & 0.682     &9     \\
 MCCT \cite{MCCT}                   & \textcolor{blue}{\underline{0.714}} & -         & 0.695     &8     \\ 
\hline
 SiamFC \cite{SiamFC}               & 0.607       & 0.516     & 0.582     &86   \\
 SA-Siam \cite{SASiam}              & 0.677       & 0.610     & 0.657     &50    \\
 RASNet \cite{RASNet}               & 0.670       & -         & 0.642     &83    \\
 SiamRPN \cite{SiamRPN}             & 0.658       & 0.592     & 0.637     &200   \\
 C-RPN \cite{CascadeSiamRPN}                & 0.675       & -         & 0.663     &23   \\
 SPM \cite{SPM}                 & 0.693       & 0.653     & 0.687     &120   \\
 SiamRPN++ \cite{SiamRPNpp}            & 0.691       & 0.662     & 0.696     &35   \\
\hline
 Meta-Tracker \cite{MetaTracker}     & 0.684       & 0.627     & 0.658     &-   \\
 MemTracker \cite{MemTracker}        & 0.642       & 0.610     & 0.626     &50    \\
 UnifiedDet \cite{UnifiedDet}        & 0.656       & -         & 0.647     &3    \\
 MLT \cite{MLT}             & 0.621       & -         & 0.611     &48   \\
 GradNet \cite{GradNet}     & 0.670       & 0.597     & 0.639     &80   \\
\hline
 MDNet \cite{MDNet}               & 0.708      & 0.645      & 0.678     &1     \\
 VITAL \cite{VITAL}               & 0.710      & 0.657      & 0.682     &2  \\
 ATOM  \cite{ATOM}               & -          & 0.628      & 0.671     &30   \\
 DiMP  \cite{DiMP}               & 0.691      & 0.654      & 0.684     &43   \\
\hline \hline
 FCOS-MAML           & \textcolor{red}{\textbf{0.714}}  & \textcolor{blue}{\underline{0.665}}  & \textcolor{blue}{\underline{0.704}}  & 42   \\ 
 Retina-MAML          & 0.709  & \textcolor{red}{\textbf{0.676}}  & \textcolor{red}{\textbf{0.712}}  & 40   \\ \hline            
\end{tabular}
\end{center}
\vspace{-5mm}
\caption{Comparison with SOTA trackers on OTB dataset. Trackers are grouped into CF-based methods, siamese-network-based methods, meta-learning-based methods, and miscellaneous. Numbers in \textcolor{red}{red} and \textcolor{blue}{\underline{blue}} are the best and the second best results, respectively.}
\label{TableOTB}
\end{table}

\subsection{Comparison with SOTA Trackers}

\begin{figure}[t!]
\centering
\begin{subfigure}{.5\linewidth}
  \centering
  \includegraphics[width=1.0\linewidth]{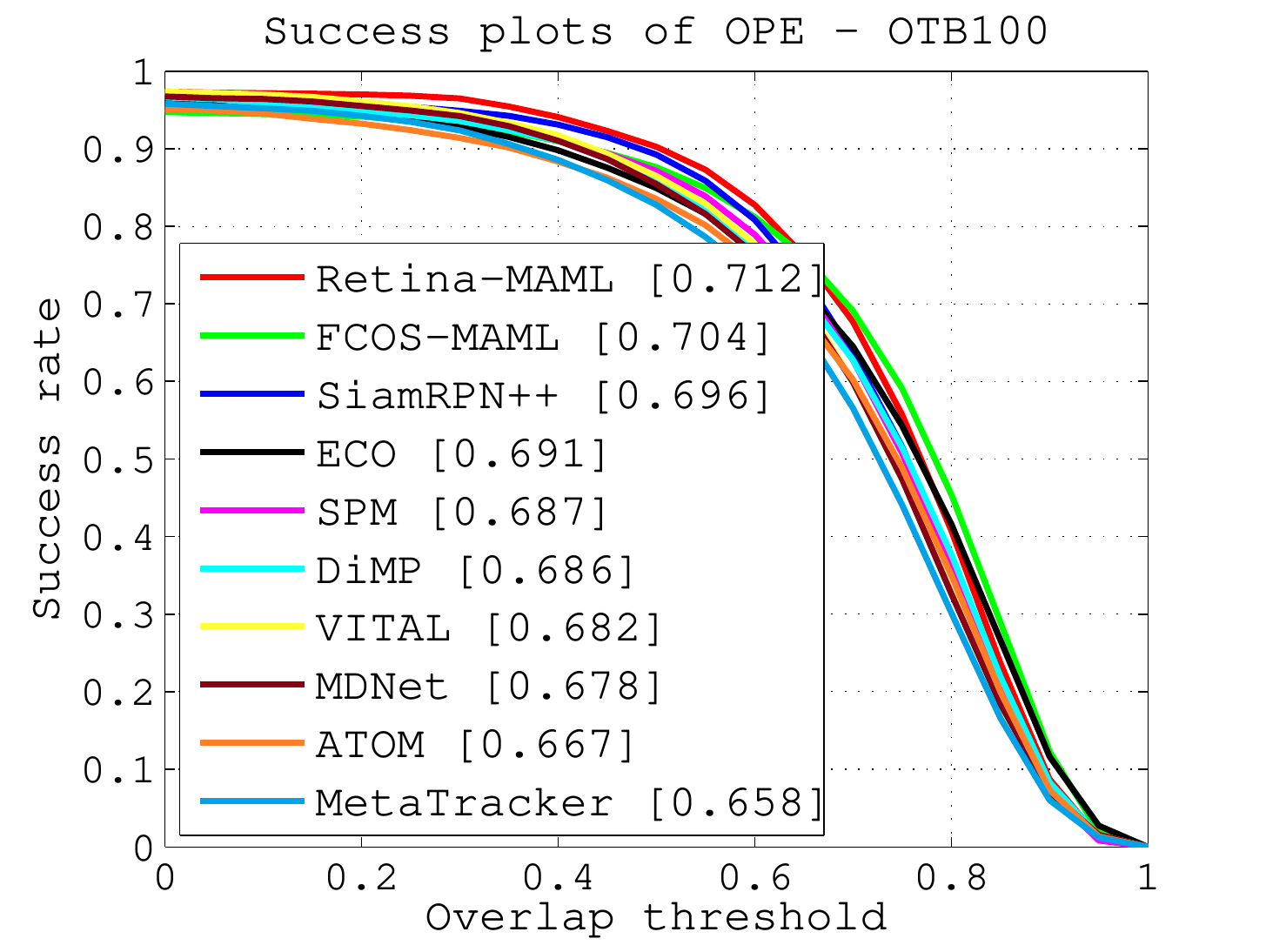}
\end{subfigure}%
\begin{subfigure}{.5\linewidth}
  \centering
  \includegraphics[width=1.0\linewidth]{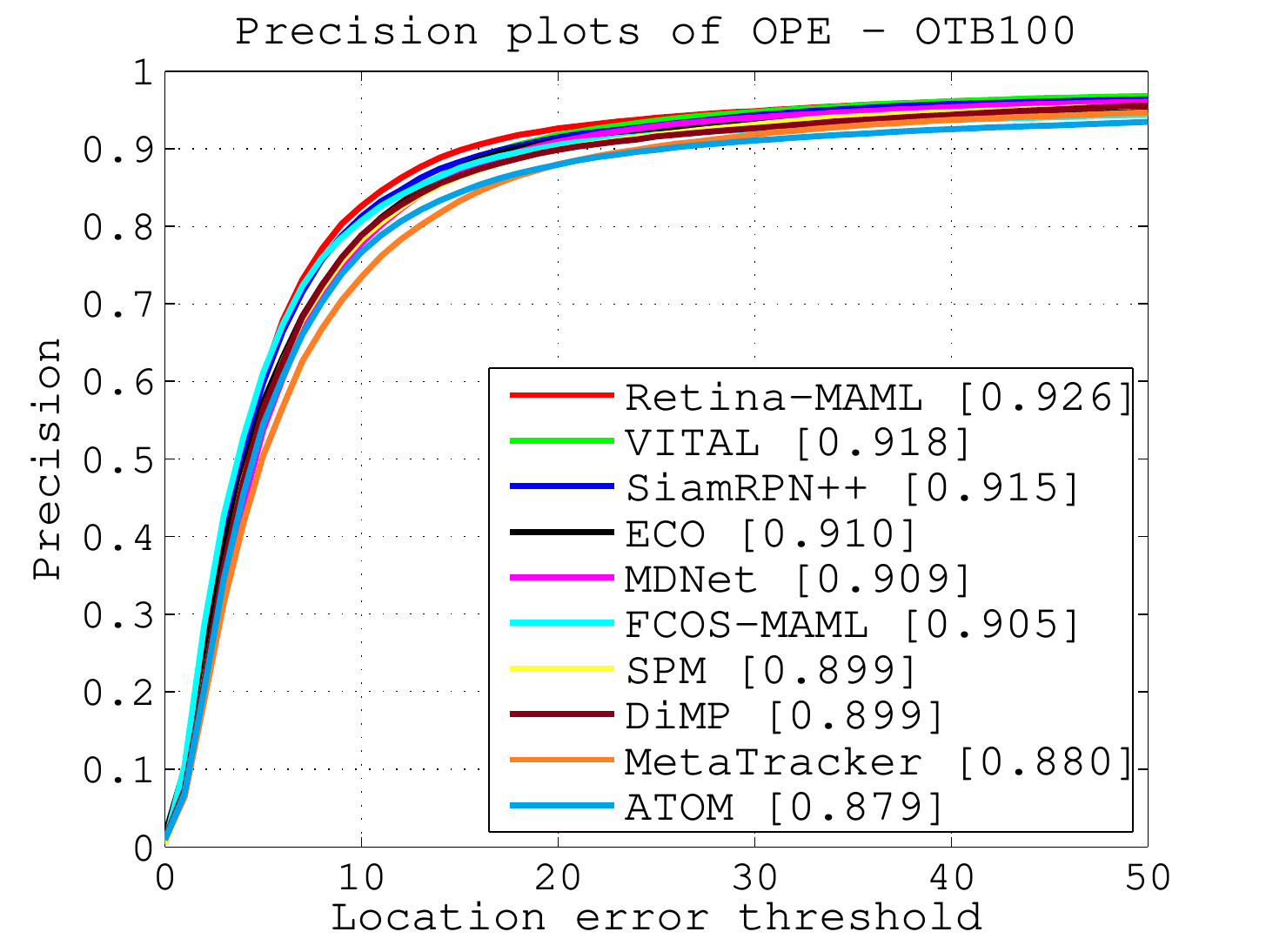}
\end{subfigure}
\caption{The success plot and precision plot on OTB-100.}
\label{FigOTBCurve}
\end{figure}

\textbf{Evaluation on OTB:} We evaluate both our trackers FCOS-MAML and Retina-MAML on OTB 2013/50/100 benchmarks \cite{OTB}. We follow the one pass evaluation (OPE) protocol, and report the AUC scores of success plot. Table \ref{TableOTB} compares our trackers with some recent top-performing trackers. On OTB-100, FCOS-MAML and RetinaNet-MAML achieve striking AUC scores of 0.704 and 0.712, respectively. To the best of our knowledge, Retina-MAML is the best-performing tracker ever on OTB. 

In this table, Meta-Tracker and UnifiedDet are two recent trackers which also use MAML to assist online training. Compared with them, our trackers achieve over 8\% relative gain in AUC and still run in real-time. For the first time, meta-learning-based methods are shown to be very competitive against the mainstream solutions. The detailed success plot and precision plot on OTB-100 are shown in Fig. \ref{FigOTBCurve}. 

\begin{table}[]
\centering
\begin{tabular}{l|ccc}
\hline
            & EAO   & Accuracy & Robustness \\ \hline
DRT \cite{DRT}       & 0.356 & 0.519    & 0.201      \\
SiamRPN++ \cite{SiamRPNpp}   & 0.414 & 0.600    & 0.234      \\
UPDT \cite{UPDT}        & 0.378 & 0.536    & 0.184      \\
LADCF \cite{LADCF}      & 0.389 & 0.503    & \textcolor{blue}{\underline{0.159}}      \\
ATOM \cite{ATOM}        & 0.401 & 0.590    & 0.204      \\
DiMP-18 \cite{DiMP}     & 0.402 & 0.594    & 0.182      \\
DiMP-50 \cite{DiMP}     & \textcolor{blue}{\underline{0.440}} & 0.597    & \textcolor{red}{\textbf{0.153}}      \\ \hline
FCOS-MAML   & 0.392 & \textcolor{red}{\textbf{0.635}}    & 0.220      \\
Retina-MAML & \textcolor{red}{\textbf{0.452}} & \textcolor{blue}{\underline{0.604}}    & \textcolor{blue}{\underline{0.159}}      \\ \hline
\end{tabular}
\caption{Comparison with SOTA trackers on VOT-2018. The backbone used in our trackers is ResNet-18. }
\label{TableVOT}
\end{table}

\textbf{Evaluation on VOT:} Our trackers are tested on the VOT-2018 benchmark \cite{VOT18} in comparison with six SOTA trackers. We follow the official evaluation protocol and adopt Expected Average Overlap (EAO), Accuracy, and Robustness as the metrics. The results are reported in Table \ref{TableVOT}. Retina-MAML achieves the top-ranked performance on EAO criteria and FCOS-MAML also shows a strong performance. Interestingly, FCOS-MAML has the highest accuracy score among all the trackers. We have observed a similar phenomenon in Fig. \ref{FigOTBCurve} for OTB dataset. FCOS-MAML gets the highest success rates when the overlap threshold is greater than 0.7. This suggests that anchor-free detectors can predict very precise bounding boxes.

\begin{table}[]
\centering
\begin{tabular}{l|cc|c}
\hline
\multirow{2}{*}{} & \multicolumn{2}{c|}{TrackingNet} & LaSOT-test \\
                  & AUC       & N-Prec.       & AUC \\ \hline
C-RPN \cite{CascadeSiamRPN}        & 0.669            & 0.746         & 0.455      \\
SiamRPN++ \cite{SiamRPNpp}        & 0.733            & 0.800         & 0.496      \\
SPM \cite{SPM}               & 0.712            & 0.779         & 0.471      \\
ATOM \cite{ATOM}              & 0.703            & 0.771         & 0.515      \\
DiMP-18 \cite{DiMP}          & 0.723            & 0.785         & \textcolor{blue}{\underline{0.532}}      \\
DiMP-50 \cite{DiMP}          & \textcolor{blue}{\underline{0.740}}            & \textcolor{blue}{\underline{0.801}}         & \textcolor{red}{\textbf{0.569}}      \\ \hline
FCOS-MAML         & \textcolor{red}{\textbf{0.757}}            & \textcolor{red}{\textbf{0.822}}         & 0.523      \\
Retina-MAML       & 0.698             & 0.786        & 0.480      \\ \hline
\end{tabular}
\vspace{-2mm}
\caption{Comparison with SOTA trackers on TrackingNet and LaSOT. We present the AUC of the success plot and and the normalized precision (N-prec.).}
\label{TabelTrackingNet}
\end{table}

\textbf{Evaluation on LaSOT and TrackingNet:} TrackingNet \cite{TrackingNet} and LaSOT \cite{LaSOT} are two large-scale datasets for visual tracking. The evaluation results on these two datasets are detailed in Table \ref{TabelTrackingNet}. Results show that FCOS-MAML performs favorably against SOTA trackers, although many of them are using a more powerful backbone ResNet-50. When compared with the recent DiMP-18 tracker which uses the same backbone network as ours, FCOS-MAML shows a significant gain on TrackingNet and a slight loss on LaSOT. We suspect that our straightforward online updating strategy may not be suitable for very long sequences which are often seen in LaSOT. 

\section{Conclusion} \label{SectionConclusion}
In this paper, we have proposed a three-step procedure to convert a general object detector into a tracker. Offline MAML training prepares the detector for quick domain adaption as well as efficient online update. The resulting instance detector is an elegant template-free tracker which fully benefits from the advancement in object detection. While the two constructed trackers achieve competitive performance against SOTA trackers in datasets with short videos, their performance on LaSOT still has room for improvement. In the future, we plan to investigate the online updating strategy for long sequences. 

\renewcommand{\thesubsection}{\Alph{subsection}}
\section*{Appendix}

\subsection{Shared blocks in backbone network}

As depicted in Fig. \ref{FigBranch}, two independent copies of block-4 in ResNet-18 are used for classification branch and regression branch, respectively. We argue that this design is not necessary but just for a more comprehensive analysis in what roles the two branches play. We also train a FCOS-MAML tracker that shares block-4 in both two branches. Note that the parameters in block-4 are still online trainable. The quantitative comparisons are shown in Table. \ref{TableSharedBlock}. The tracker that shares block-4 achieves comparable performance against its independent counterpart.

\begin{table}[h]
\centering
\begin{tabular}{c|ccc}
\hline
Shared     & OTB-100 & VOT-18 & LaSOT \\
block-4?   & (AUC)   & (EAO)  & (AUC) \\ \hline
           & 0.704   & 0.392  & 0.496 \\ \hline
\checkmark & 0.712   & 0.407  & 0.503 \\ \hline
\end{tabular}
\caption{The performance of FCOS-MAML trackers with and without shared block-4 in ResNet-18 backbone.}
\label{TableSharedBlock}
\end{table}

\subsection{Results on VOT-2019}

VOT-2019 \cite{VOT19} is a recent proposed benchmark on visual object tracking. Our two trackers are evaluated on the VOT-2019 benchmark in comparison with five SOTA trackers. The results are summarized in Table \ref{TableVOT19}. Both two trackers achieve competitive performances against SOTA trackers. Our FCOS-MAML also gets the highest accuracy among all the trackers.

\begin{table}[h]
\centering
\begin{tabular}{l|ccc}
\hline
Tracker     & EAO   & Robustness & Accuracy \\ \hline
SPM \cite{SPM}         & 0.275 & 0.507      & 0.577    \\
SiamRPN++ \cite{SiamRPNpp}  & 0.285 & 0.482      & 0.599    \\
SiamDW \cite{SiamDW}     & 0.299 & 0.467      & 0.600    \\
ATOM \cite{ATOM}        & 0.292 & 0.411      & \textcolor{blue}{\underline{0.603}}    \\
DiMP \cite{DiMP}        & \textbf{\textcolor{red}{0.379}} & \textbf{\textcolor{red}{0.278}}      & 0.594    \\ \hline
FCOS-MAML   & 0.295 & 0.421      & \textbf{\textcolor{red}{0.637}}    \\
Retina-MAML & \textcolor{blue}{\underline{0.313}} & \textcolor{blue}{\underline{0.366}}      & 0.570    \\ \hline
\end{tabular}
\caption{Comparison with SOTA trackers on VOT-2019. Numbers in \textbf{\textcolor{red}{red}} and \textcolor{blue}{\underline{blue}} are the best and the second best results, respectively.}
\label{TableVOT19}
\end{table}

\subsection{More powerful backbone}

In object detection area, it is widely acknowledged that more powerful backbones can contribute to the detection performance. We also witness the similar trends in our proposed FCOS-MAML tracker. The results are detailed in Table \ref{TableBackbone}. It clearly shows that the tracker can benefit from more powerful backbones like ResNet-50, especially on VOT-18 benchmark.

\begin{table}[h]
\centering
\setlength\tabcolsep{2.5pt}
\begin{tabular}{c|cccc}
\hline
Backbone     & OTB-100 & VOT-18 & VOT-19 & TrackingNet \\
network   & (AUC)   & (EAO)  & (EAO) & (AUC) \\ \hline
ResNet-18  & 0.704   & 0.392 & 0.295 & 0.757 \\ \hline
ResNet-50  & \textbf{0.709}   & \textbf{0.444} & \textbf{0.306} & \textbf{0.758} \\ \hline
\end{tabular}
\caption{The performance of FCOS-MAML trackers equipped with ResNet-18 and ResNet-50 backbones.}
\label{TableBackbone}
\end{table}

{\small
\bibliographystyle{ieee_fullname}
\bibliography{egbib}

\begin{thebibliography}{10}\itemsep=-1pt

\bibitem{MAML++}
Antreas Antoniou, Harrison Edwards, and Amos Storkey.
\newblock How to train your maml.
\newblock {\em arXiv preprint}, 2018.

\bibitem{SiamFC}
Luca Bertinetto, Jack Valmadre, Joao~F Henriques, Andrea Vedaldi, and Philip~HS
  Torr.
\newblock Fully-convolutional siamese networks for object tracking.
\newblock In {\em ECCV}, pages 850--865, 2016.

\bibitem{DiMP}
Goutam Bhat, Martin Danelljan, Luc Van~Gool, and Radu Timofte.
\newblock Learning discriminative model prediction for tracking.
\newblock In {\em ICCV}, 2019.

\bibitem{UPDT}
Goutam Bhat, Joakim Johnander, Martin Danelljan, Fahad Shahbaz~Khan, and
  Michael Felsberg.
\newblock Unveiling the power of deep tracking.
\newblock In {\em Proceedings of the European Conference on Computer Vision
  (ECCV)}, pages 483--498, 2018.

\bibitem{MLT}
Janghoon Choi, Junseok Kwon, and Kyoung~Mu Lee.
\newblock Deep meta learning for real-time target-aware visual tracking.
\newblock In {\em ICCV}, pages 911--920, 2019.

\bibitem{ATOM}
Martin Danelljan, Goutam Bhat, Fahad~Shahbaz Khan, and Michael Felsberg.
\newblock Atom: Accurate tracking by overlap maximization.
\newblock In {\em CVPR}, pages 4660--4669, 2019.

\bibitem{ECO}
Martin Danelljan, Goutam Bhat, Fahad Shahbaz~Khan, and Michael Felsberg.
\newblock Eco: Efficient convolution operators for tracking.
\newblock In {\em CVPR}, pages 6638--6646, 2017.

\bibitem{LaSOT}
Heng Fan, Liting Lin, Fan Yang, Peng Chu, Ge Deng, Sijia Yu, Hexin Bai, Yong
  Xu, Chunyuan Liao, and Haibin Ling.
\newblock Lasot: A high-quality benchmark for large-scale single object
  tracking.
\newblock In {\em CVPR}, pages 5374--5383, 2019.

\bibitem{CascadeSiamRPN}
Heng Fan and Haibin Ling.
\newblock Siamese cascaded region proposal networks for real-time visual
  tracking.
\newblock In {\em CVPR}, pages 7952--7961, 2019.

\bibitem{MAML}
Chelsea Finn, Pieter Abbeel, and Sergey Levine.
\newblock Model-agnostic meta-learning for fast adaptation of deep networks.
\newblock In {\em ICML}, pages 1126--1135, 2017.

\bibitem{SASiam}
Anfeng He, Chong Luo, Xinmei Tian, and Wenjun Zeng.
\newblock A twofold siamese network for real-time object tracking.
\newblock In {\em CVPR}, pages 4834--4843, 2018.

\bibitem{GOT10K}
Lianghua Huang, Xin Zhao, and Kaiqi Huang.
\newblock Got-10k: A large high-diversity benchmark for generic object tracking
  in the wild.
\newblock {\em arXiv preprint arXiv:1810.11981}, 2018.

\bibitem{UnifiedDet}
Lianghua Huang, Xin Zhao, and Kaiqi Huang.
\newblock Bridging the gap between detection and tracking: A unified approach.
\newblock In {\em ICCV}, pages 3999--4009, 2019.

\bibitem{IoUNet}
Borui Jiang, Ruixuan Luo, Jiayuan Mao, Tete Xiao, and Yuning Jiang.
\newblock Acquisition of localization confidence for accurate object detection.
\newblock In {\em ECCV}, pages 784--799, 2018.

\bibitem{RT-MDNet}
Ilchae Jung, Jeany Son, Mooyeol Baek, and Bohyung Han.
\newblock Real-time mdnet.
\newblock In {\em ECCV}, pages 83--98, 2018.

\bibitem{MetaRTT}
Ilchae Jung, Kihyun You, Hyeonwoo Noh, Minsu Cho, and Bohyung Han.
\newblock Real-time object tracking via meta-learning: Efficient model
  adaptation and one-shot channel pruning.
\newblock {\em arXiv preprint arXiv:1911.11170}, 2019.

\bibitem{BACF}
Hamed Kiani~Galoogahi, Ashton Fagg, and Simon Lucey.
\newblock Learning background-aware correlation filters for visual tracking.
\newblock In {\em ICCV}, pages 1135--1143, 2017.

\bibitem{Adam}
Diederik~P Kingma and Jimmy Ba.
\newblock Adam: A method for stochastic optimization.
\newblock {\em arXiv preprint arXiv:1412.6980}, 2014.

\bibitem{VOT18}
Matej Kristan, Ales Leonardis, Jiri Matas, Michael Felsberg, Roman Pflugfelder,
  Luka Cehovin~Zajc, Tomas Vojir, Goutam Bhat, Alan Lukezic, Abdelrahman
  Eldesokey, et~al.
\newblock The sixth visual object tracking vot2018 challenge results.
\newblock In {\em Proceedings of the European Conference on Computer Vision
  (ECCV)}, pages 0--0, 2018.

\bibitem{VOT19}
Matej Kristan, Jiri Matas, Ales Leonardis, Michael Felsberg, Roman Pflugfelder,
  Joni-Kristian Kamarainen, Luka Cehovin~Zajc, Ondrej Drbohlav, Alan Lukezic,
  Amanda Berg, et~al.
\newblock The seventh visual object tracking vot2019 challenge results.
\newblock In {\em ICCVW}, pages 0--0, 2019.

\bibitem{SiamRPNpp}
Bo Li, Wei Wu, Qiang Wang, Fangyi Zhang, Junliang Xing, and Junjie Yan.
\newblock Siamrpn++: Evolution of siamese visual tracking with very deep
  networks.
\newblock In {\em CVPR}, pages 4282--4291, 2019.

\bibitem{SiamRPN}
Bo Li, Junjie Yan, Wei Wu, Zheng Zhu, and Xiaolin Hu.
\newblock High performance visual tracking with siamese region proposal
  network.
\newblock In {\em CVPR}, pages 8971--8980, 2018.

\bibitem{GradNet}
Peixia Li, Boyu Chen, Wanli Ouyang, Dong Wang, Xiaoyun Yang, and Huchuan Lu.
\newblock Gradnet: Gradient-guided network for visual object tracking.
\newblock In {\em ICCV}, pages 6162--6171, 2019.

\bibitem{MetaSGD}
Zhenguo Li, Fengwei Zhou, Fei Chen, and Hang Li.
\newblock Meta-sgd: Learning to learn quickly for few-shot learning.
\newblock {\em arXiv preprint}, 2017.

\bibitem{RetinaNet}
Tsung-Yi Lin, Priya Goyal, Ross Girshick, Kaiming He, and Piotr Doll{\'a}r.
\newblock Focal loss for dense object detection.
\newblock In {\em ICCV}, pages 2980--2988, 2017.

\bibitem{COCO}
Tsung-Yi Lin, Michael Maire, Serge Belongie, James Hays, Pietro Perona, Deva
  Ramanan, Piotr Doll{\'a}r, and C~Lawrence Zitnick.
\newblock Microsoft coco: Common objects in context.
\newblock In {\em ECCV}, pages 740--755, 2014.

\bibitem{TrackingNet}
Matthias Muller, Adel Bibi, Silvio Giancola, Salman Alsubaihi, and Bernard
  Ghanem.
\newblock Trackingnet: A large-scale dataset and benchmark for object tracking
  in the wild.
\newblock In {\em ECCV}, pages 300--317, 2018.

\bibitem{T-CNN}
Hyeonseob Nam, Mooyeol Baek, and Bohyung Han.
\newblock Modeling and propagating cnns in a tree structure for visual
  tracking.
\newblock {\em arXiv preprint}, 2016.

\bibitem{MDNet}
Hyeonseob Nam and Bohyung Han.
\newblock Learning multi-domain convolutional neural networks for visual
  tracking.
\newblock In {\em CVPR}, pages 4293--4302, 2016.

\bibitem{MetaTracker}
Eunbyung Park and Alexander~C Berg.
\newblock Meta-tracker: Fast and robust online adaptation for visual object
  trackers.
\newblock In {\em ECCV}, pages 569--585, 2018.

\bibitem{FasterRCNN}
Shaoqing Ren, Kaiming He, Ross Girshick, and Jian Sun.
\newblock Faster r-cnn: Towards real-time object detection with region proposal
  networks.
\newblock In {\em NIPS}, pages 91--99, 2015.

\bibitem{CREST}
Yibing Song, Chao Ma, Lijun Gong, Jiawei Zhang, Rynson~WH Lau, and Ming-Hsuan
  Yang.
\newblock Crest: Convolutional residual learning for visual tracking.
\newblock In {\em ICCV}, pages 2555--2564, 2017.

\bibitem{VITAL}
Yibing Song, Chao Ma, Xiaohe Wu, Lijun Gong, Linchao Bao, Wangmeng Zuo, Chunhua
  Shen, Rynson~WH Lau, and Ming-Hsuan Yang.
\newblock Vital: Visual tracking via adversarial learning.
\newblock In {\em CVPR}, pages 8990--8999, 2018.

\bibitem{DRT}
Chong Sun, Dong Wang, Huchuan Lu, and Ming-Hsuan Yang.
\newblock Correlation tracking via joint discrimination and reliability
  learning.
\newblock In {\em CVPR}, pages 489--497, 2018.

\bibitem{FCOS}
Zhi Tian, Chunhua Shen, Hao Chen, and Tong He.
\newblock Fcos: Fully convolutional one-stage object detection.
\newblock In {\em ICCV}, 2019.

\bibitem{CFNet}
Jack Valmadre, Luca Bertinetto, Jo{\~a}o Henriques, Andrea Vedaldi, and
  Philip~HS Torr.
\newblock End-to-end representation learning for correlation filter based
  tracking.
\newblock In {\em CVPR}, pages 2805--2813, 2017.

\bibitem{SPM}
Guangting Wang, Chong Luo, Zhiwei Xiong, and Wenjun Zeng.
\newblock Spm-tracker: Series-parallel matching for real-time visual object
  tracking.
\newblock In {\em CVPR}, pages 3643--3652, 2019.

\bibitem{MCCT}
Ning Wang, Wengang Zhou, Qi Tian, Richang Hong, Meng Wang, and Houqiang Li.
\newblock Multi-cue correlation filters for robust visual tracking.
\newblock In {\em CVPR}, pages 4844--4853, 2018.

\bibitem{RASNet}
Qiang Wang, Zhu Teng, Junliang Xing, Jin Gao, Weiming Hu, and Stephen Maybank.
\newblock Learning attentions: residual attentional siamese network for high
  performance online visual tracking.
\newblock In {\em CVPR}, pages 4854--4863, 2018.

\bibitem{OTB}
Yi Wu, Jongwoo Lim, and Ming-Hsuan Yang.
\newblock Object tracking benchmark.
\newblock {\em T-PAMI}, 37(9):1834--1848, 2015.

\bibitem{LADCF}
Tianyang Xu, Zhen-Hua Feng, Xiao-Jun Wu, and Josef Kittler.
\newblock Learning adaptive discriminative correlation filters via temporal
  consistency preserving spatial feature selection for robust visual object
  tracking.
\newblock {\em TIP}, 2019.

\bibitem{MemTracker}
Tianyu Yang and Antoni~B Chan.
\newblock Learning dynamic memory networks for object tracking.
\newblock In {\em PECCV}, pages 152--167, 2018.

\bibitem{RTINet}
Yingjie Yao, Xiaohe Wu, Lei Zhang, Shiguang Shan, and Wangmeng Zuo.
\newblock Joint representation and truncated inference learning for correlation
  filter based tracking.
\newblock In {\em ECCV}, pages 552--567, 2018.

\bibitem{SiamDW}
Zhipeng Zhang and Houwen Peng.
\newblock Deeper and wider siamese networks for real-time visual tracking.
\newblock In {\em CVPR}, pages 4591--4600, 2019.

\bibitem{DaSiamRPN}
Zheng Zhu, Qiang Wang, Bo Li, Wei Wu, Junjie Yan, and Weiming Hu.
\newblock Distractor-aware siamese networks for visual object tracking.
\newblock In {\em ECCV}, pages 101--117, 2018.

\end{thebibliography}
}

\end{document}